\pdfoutput=1
\documentclass[11pt]{article}

\usepackage[final]{acl}

\usepackage{inconsolata}

\usepackage{graphicx}
\usepackage{subcaption}
\usepackage{float}
\usepackage[justification=justified]{caption}	% makes captions ragged right - thanks to Bryce Lobdell
\usepackage{lscape}                                         % Useful for wide tables or figures.
\usepackage{wrapfig}
\usepackage[lined,ruled,linesnumbered]{algorithm2e}
\usepackage{animate} %% convert frames to video

\usepackage{booktabs}                   % Publication quality tables
\usepackage{multirow}

\usepackage{makecell}

\usepackage{paralist}
\usepackage{enumitem}

\usepackage{bm}                          % Make bold, italic math symbols
\usepackage{epsfig}                      % for figures
\usepackage{graphicx}                  % another package that works for figures
\usepackage{times}
\usepackage{latexsym}               % from emnlp 2024 template (ACL)

\usepackage{mathptmx}
\usepackage{mathtools}
\usepackage{amssymb,amsmath}   % Short math guide for LaTeX ftp://ftp.ams.org/pub/tex/doc/amsmath/short-math-guide.pdf

\usepackage{units}
\usepackage{color}
\usepackage[T1]{fontenc}    % use 8-bit T1 fonts
\usepackage{amsfonts}       % blackboard math symbols
\usepackage[utf8]{inputenc} % allow utf-8 input
\usepackage{nicefrac}       % compact symbols for 1/2, etc.
\usepackage{microtype}      % microtypography
\usepackage{comment}

\usepackage{inconsolata}

\usepackage{url}  % Hyphenation of URLs.
\DeclareMathOperator*{\argmax}{\arg\!\max}

\newlength\paramargin
\newlength\figmargin

\newlength\secmargin
\newlength\figcapmargin
\newlength\tabcapmargin

\newcommand{\figcaption}[2]
{
\caption{
\textbf{#1.}  % Figure caption title
#2            % Caption
}
}

\newcommand{\secref}[1]{Section~\ref{sec:#1}}
\newcommand{\figref}[1]{Figure~\ref{fig:#1}} 
\newcommand{\tabref}[1]{Table~\ref{tab:#1}}
\newcommand{\eqnref}[1]{\eqref{eq:#1}}

\newcommand{\appref}[1]{Appendix~\ref{app:#1}}  % added by hg

\long\def\ignorethis#1{}

\newcommand{\tb}[1]{\textbf{#1}}

\newbox\jsavebox%

\newcommand{\best}[1]{\textbf{#1}}
\newcommand{\second}[1]{\underline{#1}}

\newcommand{\mlp}{\texttt{MLP}}
\def\xi{\mathbf{x}_i}

\def\ours{EI-VLG}
\def\lfvlg{LFVLG}
\def\capgenllavaimage{LV-Img-34B}
\def\capgenllavavideo{LV-Vid-7B}
\def\capgenvideorecap{VideoRecap}
\def\ourtextenc{EE}

\def\x{\mathbf{x}}

\graphicspath{{figure}, {example}}

\title{Infusing Environmental Captions for Long-Form Video Language Grounding}
\newcommand*\samethanks[1][\value{footnote}]{\footnotemark[#1]}

\author{
Hyogun Lee\thanks{Equal contributor},
Soyeon Hong\samethanks[1],
Mujeen Sung\thanks{Corresponding author},
Jinwoo Choi\samethanks[2] \\
Kyung Hee University \\
{\tt\small \{gunsbrother,soyeonhong,mujeensung,jinwoochoi\}@khu.ac.kr}
}

\begin{document}
\maketitle
% \begin{abstract}
% This document is a supplement to the general instructions for *ACL authors. It contains instructions for using the \LaTeX{} style files for ACL conferences.
% The document itself conforms to its own specifications, and is therefore an example of what your manuscript should look like.
% These instructions should be used both for papers submitted for review and for final versions of accepted papers.
% \end{abstract}
\begin{abstract}

In this work, we tackle the problem of long-form video-language grounding (VLG).
Given a long-form video and a natural language query, a model should temporally localize the precise moment that answers the query.
Humans can easily solve VLG tasks, even with arbitrarily long videos, by discarding irrelevant moments using extensive and robust knowledge gained from experience.
Unlike humans, existing VLG methods are prone to fall into superficial cues learned from small-scale datasets, even when they are within irrelevant frames.
To overcome this challenge, we propose {\ours}, a VLG method that leverages richer textual information provided by a Multi-modal Large Language Model (MLLM) as a proxy for human experiences, helping to effectively exclude irrelevant frames.
We validate the effectiveness of the proposed method via extensive experiments on a challenging EgoNLQ benchmark.
%
% {\ours} achieves state-of-the-art performance on EgoNLQ with an R5@0.3 of 35.2\%.
% Tone downed a bit.
%
% We will release the training code, model weights and the environmental descriptions upon acceptance.

\end{abstract}

\begin{figure}[t]
\centering
    \includegraphics[width=\linewidth]{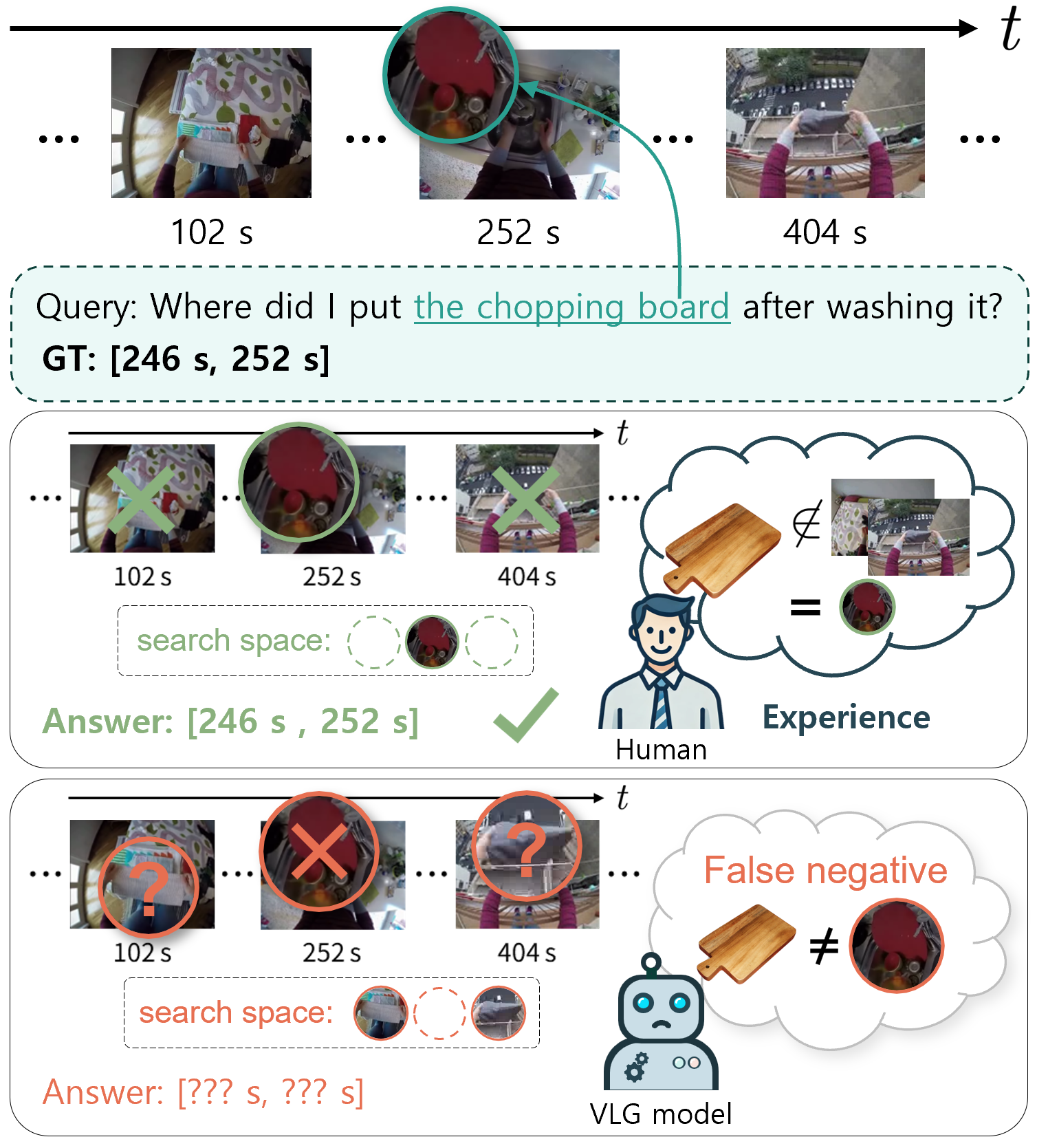}
    %Teaser figure. What are the input and output of the method?
    %
    % 비디오가 아무리 길어도 사람은 금방 찾는다
    % 사람은 질문과 가장 관련 있는 장면을 연상하고 매치하면서 search space를 줄여 효율적으로 찾는다
    % 하지만 AI model는 장면이 아닌 오브젝트에 집중하기에 비디오가 길어지면 false negatives가 늘어난다
    % (원래 false positives 였는데 "search space 줄인답시고 멍청한 모델에게 함부로 관련이 적다고 생각하는 부분을 제거하면 안 된다"를 전달하기 위해 false negatives로 변경)
    % \figcaption{Humans are effective \lfvlg{} solvers}{
    % No matter how long a video is, humans can quickly find relevant information. Humans reduce the \textit{search space} efficiently by recalling and matching scenes most related to the query. However, AI models tend to focus on objects rather than scenes, leading to an increase in false negatives as the video length increases.
    % }
    %
    % Before revision by Jinwoo
    % \figcaption{Humans are effective \lfvlg{} solvers}{
    % \hg{TODO/ though looking weird, it won't be hard for humans to find the real chopping board.
    % Existing VLG models discard the GT moment merely because it does not have wooden texture.
    % }
    % No matter how long a video is, humans can quickly find relevant information.
    % Humans reduce the \textit{search space} by recalling and matching scenes most related to the query.
    % However, AI models tend to focus on objects rather than scenes, leading to an increase in false negatives as the video length increases.
    % \hg{can skin color cause an ethical issue?}
    % }

    % After revision by Jinwoo
    \captionof{figure}{
    \textbf{How do humans and machines solve the long-form video-language grounding problem?}
    The example illustrates how humans can easily localize the red chopping board using extensive and robust knowledge gained from experience.
    %
    % Humans utilize extensive and robust knowledge gained from experience, such as knowing that a chopping board is likely to be in the \emph{kitchen} rather than the bedroom or the balcony, to effectively reduce the search space.
    %
    In contrast, VLG models trained on small-scale datasets might incorrectly discard the ground truth moment because the chopping board does not have a wooden texture.
    %
    % Inspired by human capabilities of search space reduction using environmental cues, we propose leveraging the extensive knowledge of a multi-modal large language model to address the long-form video-language grounding problem.
    % \hg{need to address: more human/robot-like}
    }
    \label{fig:teaser}
\end{figure}

\section{Introduction}
\label{sec:intro}

Given an arbitrarily long video, humans can easily localize moments of interest. 
For example, humans can quickly identify the precise moment the camera-wearer \emph{puts a chopping board on the kitchen sink}, as illustrated in \figref{teaser}.
Humans quickly discard irrelevant moments such as moments involving \emph{folding the laundry}, by leveraging extensive and robust knowledge gained from experience.
We aim to develop a model that mimics this human-like capability of reducing the search space to solve the long-form video-language grounding (\lfvlg{}) problem.

In the \lfvlg{} problem, a model should temporally localize the specific moment within a long-form video that answers a given natural language query.
% \mj{`language query' -> `natural language query' or `text query'}
%
Developing a high-performance \lfvlg{} model could significantly benefit many high-impact applications, such as content-based video search, augmented reality, and video editing.

\begin{figure*}[t]
\includegraphics[width=\linewidth]{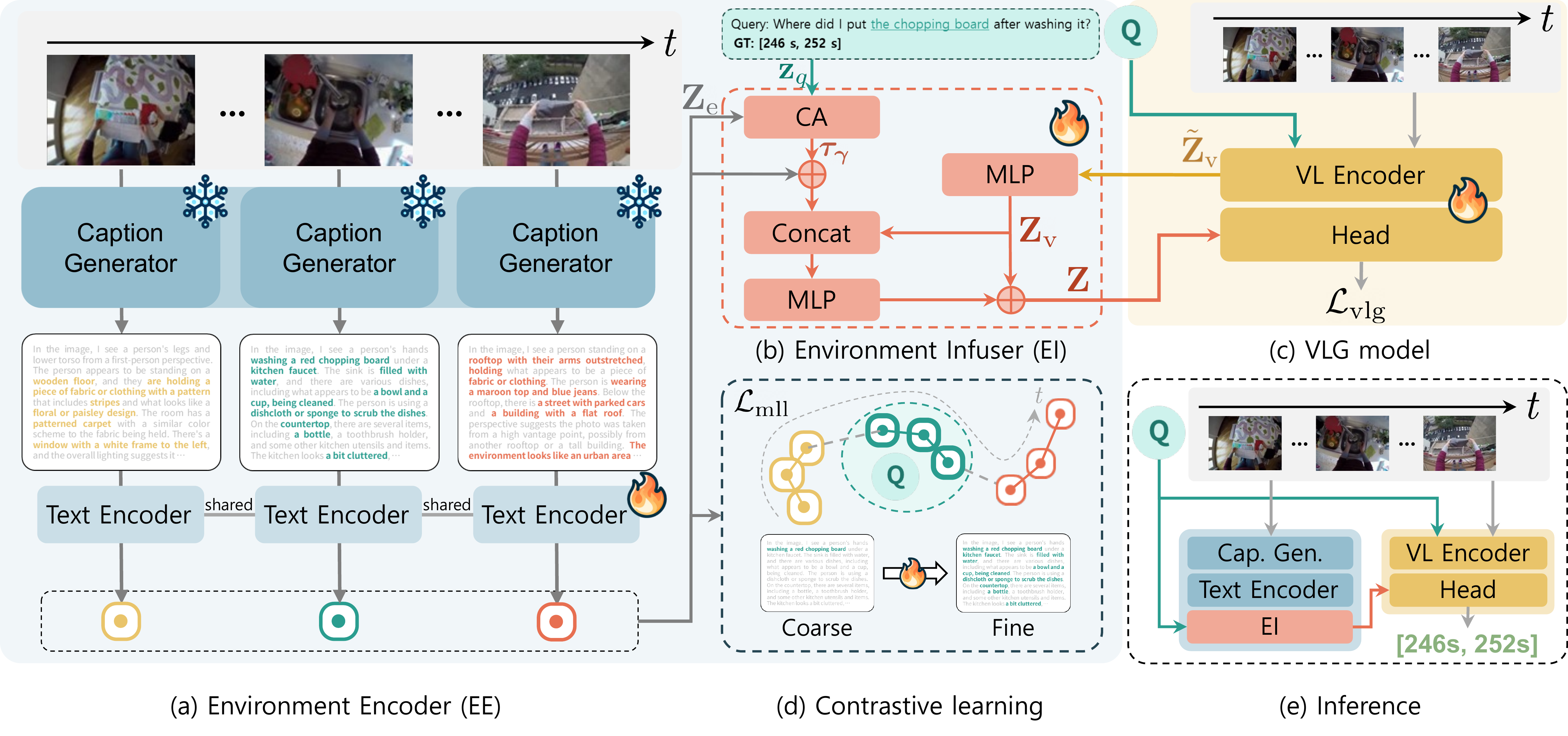}
\figcaption{Overview of \ours{}}{
\ours{} consists of three components: (a) environment encoder (\secref{ee}),
(b) environment infuser (\secref{ei}), and
(c) video-language model (\secref{vlg}).
(d) We fine-tune the environment encoder to encourage the encoded environment feature vectors to be suitable for attention with query embedding.
(e) During inference, \ours{} effectively reduces the search space by infusing the environment knowledge.
}
\label{fig:overview}
\end{figure*}

%
% After revision by Jinwoo
Compared to short-form VLG, where an input video is rather short, \lfvlg{} is much more challenging.
On the Charades-STA~\cite{Gao_2017_ICCV}, a widely used short-form VLG benchmark, the current state-of-the-art performance is R5@0.5 score of 91.94\%~\cite{zeng2024unimd}.
In contrast, on the EgoNLQ~\cite{grauman2022ego4d}, a recent \lfvlg{} benchmark, the state-of-the-art performance is only
% R5@0.3 of 34.29\%~\cite{di2023groundvqa}.
% \hg{(agree with that) 34.29\% in R5@0.3 and 23.37\% in R5@0.5~\cite{di2023groundvqa}}
34.3\% in R5@0.3 and 23.4\% in R5@0.5~\cite{di2023groundvqa}.
\lfvlg{} is challenging because it is a \emph{needle-in-a-haystack} problem.
%
% \jw{Write some supporting evidence here.}
% Before revision by Jinwoo
% \hg{
% The average proportion of ground truth moments relative to the entire video duration--referred to as coverage--is merely one-tenth that of short-form datasets.
% For instance, Charades-STA~\cite{Gao_2017_ICCV} has a coverage of 27.0\%., whereas EgoNLQ~\cite{grauman2022ego4d} has only 2.3\%.
% This performance gap is understandable because, to achieve the same coverage, we need to successfully filter out 90\% of the video as irrelevant parts.
% }  
% AFter revision by Jinwoo
The average proportion of ground truth moments to the entire video duration--referred to as GT coverage--  of the \lfvlg{} task is an order of magnitude smaller than the GT coverage of short-form VLG tasks.
For instance, Charades-STA has a GT coverage of 27.0\%, whereas EgoNLQ has a GT coverage of 2.3\% only.
This substantial difference in GT coverage makes \lfvlg{} particularly challenging, as it requires accurately discarding approximately 90\% of the video as irrelevant moments.

Inspired by human capabilities of search space reduction using environment cues, we introduce a novel approach to address {\lfvlg}: environment infusion for video-language grounding ({\ours}).
We leverage the extensive knowledge of a multi-modal large language model (MLLM)~\cite{li2023llavamed, liu2024llavanext, liu2023improvedllava, zhang-etal-2023-video, lin2023video, zhang2023llava, li2022blip, li2023blip2, instructblip, ren2023timechat, maaz2023videochatgpt} to effectively reduce search space.
Given an input video, we generate captions at regular short-term intervals and encode them using a text encoder to serve as environment cues.
Despite being zero-shot, MLLM-generated captions provide much more detailed contextual descriptions than a dataset-specific captioner~\cite{islam2024video}, as shown in \figref{llava}.
%
% We encode the environmental cues using a text encoder.
%
% The environment infuser comprises a text encoder and an injection module, both fine-tuned for grounding. \jw{should be updated.}
%
We then infuse these environment cues into a VLG model.
By leveraging these cues, {\ours} can capture details and distinguish fine-grained differences among moments within an input video.
We validate the proposed method on a challenging EgoNLQ benchmark through extensive experiments.
The proposed method shows favorable performance compared to the state-of-the-art.

%
% 우리의 솔루션
% itemize 하니까 위의 Existing approaches to video-language grounding (VLG) are categorized into three types
% 랑 겹쳐, 난잡해보여서 줄글로 했습니다. -> itemize로 수정 완료
% \jw{Use itemize for the contribution summary.}
%
% LTVLG에서 숏컷을 알아내고 활용하는 법=추출+합침을 알아냈다. ours는 기존 VLG 모델에 붙일 수 있다.
% 다양한 실험으로 구조적인 search space를 줄였다.
% 실험으로 뭐뭐 달성 없으면 "코드 풀 거다"

% Before revision by Jinwoo
% We introduce a novel approach to solving {\lfvlg} called {\ours}. The key contributions of our work are 3-fold:
% \begin{itemize}
%     \item We demonstrate that MLLMs can provide empirical information used by humans to solve {\lfvlg} and find how this information can be infused into existing VLG models.
%     \item We prove that MLLMs are suitable for providing rich environmental information. We measure the faithfulness of MLLMs in the light of the objects they describe. 
%     \item We will make the description data, training code, and model weights publicly available.
% \end{itemize}
%
% After revision by Jinwoo
To summarize, we make the following key contributions.
\begin{itemize}
    % \item Inspired by human search space reduction capabilities, we propose a novel \lfvlg{} method using environment cues to effectively reduce search space.
    \item Inspired by human search space reduction capabilities, we propose {\ours}, using environment cues to effectively reduce search space.
    
    \item We validate \ours{} via thorough experiments on the challenging benchmark: EgoNLQ. 
\end{itemize}
We will release the description data, code and model weights upon acceptance.

\section{\ours}
\label{sec:method}

We introduce {\ours}, a novel \lfvlg{} method inspired by human capabilities of search space reduction using environment cues.
As illustrated in \figref{overview}, {\ours} consists of three components: i) environment encoder (EE), ii) video-language grounding model (VLG), and iii) environment infuser (EI).
To reduce the search space for \lfvlg{}, it is crucial to infuse environment cues extracted by EE into the VLG model.

Given an input video, EE extracts rich captions for short-term intervals and encodes them using a learnable text encoder. 
Then we infuse the environment cues into a VLG model.
In the following subsections, we provide detailed descriptions of each component.

% Before revision by Jinwoo
% \subsection{The EnvExt architecture}\label{sec:envext}
% % The EnvExt has 2 sub-modules: an MLLM which provides rich description of environment and a text encoder encoding it for grounding.
% % To make the text encoder attend to helpful information for grounding, we fine-tune the text encoder in a contrastive way.
% \hg{
% EnvExt consists of an MLLM and a text model.
% The MLLM takes a sub-clip as input and generates a description, which is then encoded by the text model.
% To obtain an environment feature that effectively reduces the search space in long-form scenarios, two key conditions must be met:
% (i) the captioner must generate fine-grained and rich descriptions, and
% (ii) the text encoder must understand the subtle differences between in-context captions. 
% After explaining each component, we will describe how to obtain the final environment feature.
% }

% After revision by Jinwoo
\subsection{Environment Encoder}
\label{sec:ee}
The environment encoder (EE) comprises a frozen environment caption generator, $f(\cdot)$, and a text encoder, $g(\cdot;\theta)$ with learnable parameters $\theta$. 

\paragraph{Caption generator.}
Given an $M$-frame long video, we subsample $N \ll M$ frames with the same interval to obtain $\mathbf{X}=\{\x_1, \x_2, \hdots, \x_N\}$.
% $\mathbb{V}=\{\v_i \mid 1 \leq i \leq N\}$.
%
As our caption generator, we employ an off-the-shelf MLLM, LLaVA (34B)~\cite{liu2023llava, liu2023improvedllava, liu2024llavanext}.
We empirically find that using a sufficiently large model is crucial to provide a fine-grained and rich context to a VLG model for effective search space reduction.
Then we get frame-level environmental captions $f(\mathbf{X})$.

\paragraph{Text encoder.}
We encode environmental captions using a text encoder as follows:
\begin{equation}
    \mathbf{Z_\text{e}} = g(f(\mathbf{X});\theta) \in \mathbb{R}^{N \times D_\text{t}},
\end{equation}
where $D_\text{t}$ is the feature dimension.
We also encode the textual query $\mathbf{q}$ using the same text encoder to obtain a query embedding: $\mathbf{z}_q=g(\mathbf{q}) \in \mathbb{R}^{D_\text{t}}$.

\paragraph{Text encoder learning.}
We aim for the encoded environment feature vectors to be suitable for attention with the query embedding. 
 To achieve this, we fine-tune an off-the-shelf text encoder with a contrastive learning objective.
The similarity between a query and captions within the ground truth (GT) interval should be greater than the similarity between the query and captions outside the GT interval.
Therefore, we employ the marginal log-likelihood loss~\cite{lee2020learning, min2019discrete, lee2019latent} to fully utilize multiple positive pairs given the GT interval.
Given a GT interval with start and end frame indices $s$ and $e$, we define the marginal log-likelihood loss as follows:
\begin{align}     
    % L_\text{mll}(\theta) &= -\sum_{s \leq  i \leq e} \log \text{Softmax} \left( \textbf{z}_i^{\top}\textbf{z}_q \right),
    L_\text{mll}(\theta) &= - \log \frac{\sum_{s \leq  i \leq e} \exp(\textbf{z}_i^{\top}\textbf{z}_q)}{\sum_{j=1}^N  \exp(\textbf{z}_j^{\top}\textbf{z}_q)},
    \label{eq:mll}
\end{align}
where $\mathbf{z}_i \in \mathbb{R}^{D_t}$ is the $i$-th vector of $\mathbf{Z_\text{e}}$.

\subsection{Video-Language Grounding Model}
\label{sec:vlg}
% \jw{Please dump some information here.}
% Before revision by Jinwoo
% \hg{
% The model consists of a VL Encoder and a head.
% The VL Encoder takes an $M$-frame long video $\mathbf{X}' = \{\mathbf{x}_1,\mathbf{x}_2, \dots, \mathbf{x}_M \}$ and a natural language query $\mathbf{q}$ to generate video and query features $\mathbf{Z}'_\text{v} \in \mathbb{R}^{M \times D_\text{v}}, \mathbf{Z}'_q \in \mathbb{R}^{L \times D_\text{v}}$, where $L$ is the number of query tokens and $D_\text{v}$ the feature dimension.
% There are two options for the encoder modality:
% a single multi-modal encoder or two separate encoders, one for video and one for the query.
% The head then uses $\mathbf{Z}'_\text{v}$ to generate the start and end frame indices $\hat{s}, \hat{e} \in \{1, 2, \dots, M\}$ of the interval most relevant to the query.
% Instead of using $\mathbf{Z}'_\text{v}$ directly, {\ours} incorporates environmental information by combining the features in the Environment Infuser (EI), resulting in the feature $\mathbf{Z}$ that is then fed into the head.
% <<DONE>>
% }

% After revision by Jinwoo
We can employ any existing VLG model that consists of a vision-language (VL) encoder and a temporal localization head.
The VL encoder takes an $M$-frame long video $\mathbf{\tilde{X}} = \{\mathbf{x}_1,\mathbf{x}_2, \dots, \mathbf{x}_M \}$ and a natural language query $\mathbf{q}$ to generate video and query features $\mathbf{\tilde{Z}}_\text{v} \in \mathbb{R}^{M \times D_\text{v}}, \mathbf{\tilde{Z}}_q \in \mathbb{R}^{L \times D_\text{v}}$, where $L$ is the number of query tokens and $D_\text{v}$ the feature dimension.
The temporal localization head then takes $\mathbf{\tilde{Z}}_\text{v}$ to localizes the start and end frame indices $\hat{s}, \hat{e} \in \{1, 2, \dots, M\}$ of the interval most relevant to the query.

\subsection{Environment Infuser}
\label{sec:ei}
The environment infuser (EI) enhances a VLG model's understanding of environmental information.
We infuse the environment feature vectors $\mathbf{Z}_\text{e}$ with the video feature vectors $\mathbf{Z}_\text{v}=\mlp{(\mathbf{\tilde{Z}})} \in \mathbb{R}^{M \times D_\text{v}}$ as follows:
\begin{equation}
    \mathbf{Z} = \mathbf{Z}_\text{v} + [\mathbf{Z}_\text{e} + \tau_\gamma \cdot \text{CA}(\mathbf{z}_q, \mathbf{Z}_\text{e})| \mathbf{Z}_\text{v}] \mathbf{W}.
\end{equation}
% \hg{changed ; to |, since ; is row-wise concat op.}
Here, $\text{CA}(\mathbf{z}_q, \mathbf{Z}_\text{e})$ denotes a cross-attention layer between the query embedding $\mathbf{z}_q$ and the environment feature vectors $\mathbf{Z}_\text{e}$, $|$ is the row-wise concatenation, $\tau_\gamma$ is a hyperbolic tangent function with learnable hyperparameter $\gamma$, and $\mathbf{W} \in \mathbb{R}^{2 D_\text{v} \times D_\text{v}}$ is a learnable projection matrix.
% \hg{$\gamma$ learnable, inspired by gligen https://arxiv.org/abs/2301.07093}

% problem definition 간단히
    % VLG task is to find a temporal interval $(t_s, t_e)$ given a video $v$ as a sequence of RGB frames \{v_t\}_{t=1}^T and a natural language query $q$.
    % that can answer the query

% Structure
% 비디오 모델 = 인코더 + 헤드
% Env ext = 라바 + 텍스트 모델, 라바의 exp를 env로 바꿔줌.
% infuser = 인코더 output + env ext output
% loss_env = env ext output에 multipos loss (성무진 교수님 페이퍼 사이트)
    % 의미
    % Query와 correlate: grounding task에 필요한 정보를 뽑는 법을 배움. 라바의 exp를 env로 바꿀 수 있게 됨.
    % Temporal modeling: env ext에는 input이 per-clip으로 들어가지만 여기서 temporal modeling을 하고 dense(in-video discriminative) 한 representation을 얻음

% captioner
    % 34B는 10초에 하나, 7B랑 videorecap은 2초에 하나 뽑음
    % 캡션들을 text인코더로 인코딩

% contrastive
    % dataset

% infuser 

% training

% 왜 라바 캡션 쭉쭉 이어붙여서 instruction tune 하면 안 됨?

\section{Experiments}
\label{sec:results}
% Three practices that are common in the strongest empirical papers are 
% - error analysis, 
% - ablation studies, and 
% - robustness checks
% (to e.g. choice of hyper-parameters, as well as ideally to choice of dataset)

% Before revision by Jinwoo
% We conduct all the experiments on EgoNLQ dataset. We use top-k recall as a evaluation metric

% After revision by Jinwoo
In this section, we present the experimental results to validate the proposed method.
We evaluate the proposed method on the challenging EgoNLQ~\cite{grauman2022ego4d} dataset, which consists of 14K training samples and 4K validation samples, with an average video length of 8 minutes.
%
% For comprehensive details on the dataset, please check \appref{about_egonlq_dataset}.
%
We use the top-k recall at an intersection over union (IoU) threshold value, R<k>@<threshold>, as an evaluation metric denoted.
Following the prior works~\cite{zhang-etal-2020-span, nagarajan2023egoenv, linegocentric, Pramanick_2023_ICCV, di2023groundvqa}, we report R1@0.3, R5@0.3, R1@0.5, and R5@0.5.
Please note that we do not compare with methods training a model on external datasets~\cite{ramakrishnan2023naq, Gao_2017_ICCV, ZhXuCoCVPR18, lei2021detecting}. % \jw{please cite more here.}.
We describe the comprehensive details on implementation and experimental setup in \appref{implementational_detail}.
Please note that all experiments in this section, including ablations, are single runs.
%
% \jw{Please put them in the appendix accordingly. We should say that we get the GVQA performance by running their code.}
% \mj{please refer to the specific Appendix section (e.g., Appendix A).}
% \sy{Addressed.}
% \sy{Please refer to the \appref{implementational_detail}}
% Before revision by Jinwoo
% \subsection{Quantitative evaluation}
% In this section, We compare the performance of our model with the current state-of-the-art model, on the EgoNLQ task. We show the results in  Table~\ref{tab:main_all}, {\ours} achieves the highest performance of an R5@0.3 score of 35.2\%, outperforming existing models GroundVQA~\cite{di2023groundvqa} that achieves R5@0.3 score of 34.3\%. This demonstrates the effectiveness of our model architecture and environment cues.
% \hg{(why R5@0.3 is our primary metric) }

% vs. GVQA

\begin{table}[t]
\centering

\resizebox{\columnwidth}{!}{
\begin{tabular}{l c cccc}
\toprule
\textbf{Method} & \textbf{R1@0.3} & \textbf{R5@0.3} & \textbf{R1@0.5} & \textbf{R5@0.5}
\\
\midrule
VSLNet~\cite{zhang-etal-2020-span}  % https://arxiv.org/pdf/2207.11365
& 5.5 & - & 3.1 & - \\
EgoEnv~\cite{nagarajan2023egoenv}  % https://arxiv.org/pdf/2207.11365
& 6.0 & - & 3.5 & - \\
EgoVLP~\cite{linegocentric}  % https://arxiv.org/pdf/2307.05463
& 10.8 & 18.8 & 6.8 & 13.5 \\
EgoVLPv2~\cite{Pramanick_2023_ICCV}  % https://arxiv.org/pdf/2307.05463
& 13.0 & 23.8 & 7.9 & 16.1 \\
CONE~\cite{hou2023cone}
& 14.2 & 30.3 & 8.2 & 18.0 \\
GroundVQA~\cite{di2023groundvqa}
&\best{15.3} & \second{34.3} & \second{9.4} & \second{23.4} \\
{\ours~(Ours)}  % 버전: ???
& \second{15.2} & \best{35.2} & \best{10.0} & \best{23.8} \\
\bottomrule
\end{tabular}
}
\caption{\tb{Main results on the EgoNLQ validation set.}}
\label{tab:main_all}
\end{table}

% After revision by Jinwoo
\subsection{Main Results}
We compare the performance of the proposed method with the current state-of-the-art methods in \tabref{main_all}. 
{\ours} shows the best performance 3 out of 4 metrics and on par with state-of-the-art in terms of R1@0.3.
The results validate the effectiveness of using environment cues provided by an MLLM as a proxy for human experiences.

\begin{table}[t]
\centering
\resizebox{\columnwidth}{!}{
\begin{tabular}{l c cccc}
\toprule
& \textbf{R1@0.3} & \textbf{R5@0.3} & \textbf{R1@0.5} & \textbf{R5@0.5}
\\
\midrule
{\ours} (Ours)  & 15.2 & \textbf{35.2} & \textbf{10.0} & \textbf{23.8} \\  % 버전: 102720
% \;w/ VideoRecap\cite{islam2024video} & 15.0 & 34.5 & 9.6 & 23.5 \\
\;\capgenllavaimage $\rightarrow$ $\emptyset$ & 15.3 & 34.3 & 9.4 & 23.4 \\
\;\capgenllavaimage $\rightarrow$ \capgenvideorecap & 15.0 & 34.5 & 9.6 & 23.5 \\
% \;w/ LLaVA video 7B\cite{liu2024llavanext} & 15.4 & 34.5 & 10.0 & 23.6 \\
\;\capgenllavaimage $\rightarrow$ \capgenllavavideo & \textbf{15.4} & 34.5 & \textbf{10.0} & 23.6 \\
\;Concat. $\rightarrow$ Add & 14.9 & 34.6 & 9.9 & 23.5 \\
\;Concat. $\rightarrow$ CA & 15.2 & 34.4 & 9.8 & 23.3 \\ 
\;SBERT $\rightarrow$ EgoVLP & 15.3 & 33.6 & 9.8 & 23.2 \\ 
\bottomrule
\end{tabular}
}
\caption{\tb{Ablation study}: environment infusion.}
\label{tab:abl_arch}
\end{table}

% After revision by Jinwoo
\subsection{Ablation Study}
To validate the efficacy of the proposed method, we conduct thorough ablation experiments on environment infusion, quality of environment cues, and environment encoder.
% and present the results in \tabref{abl_arch}, \tabref{abl_env_cues}, and \tabref{abl_text_encoder}.

% \subsubsection{Environment Cues, Caption Quality, and Infuser Architecture \mj{Is there a term that can include all of these?} }
\subsubsection{Environment Infusion}

% \paragraph{Effect of environment cues.}
% \hg{In the second row of \tabref{abl_arch}, we can see that }
% and present the results in \tabref{abl_arch}, \tabref{abl_env_cues}, and \tabref{abl_text_encoder}.
\tabref{abl_arch} describes the ablation study of different environment cues and infusion architecture.
The proposed method (\ours{}) outperforms the baseline without environment cues ($\emptyset$) in 3 out of 4 metrics and achieves comparable performance in terms of R1@0.3. 
The results validate the effectiveness of incorporating environment cues.

% \paragraph{Effect of caption quality.} 
To study the effect of caption quality, we compare three caption generators. 
%
% i) VideoRecap\cite{islam2024video} is a small model but it is trained on Ego4D. 
i) VideoRecap~\cite{islam2024video} is a small model but it is trained on Ego4D. 
%
% ii) \texttt{LLaVA-Video-7B-Env}~\cite{liu2024llavanext} is a multi-modal large language model (MLLM) trained on instruction tuning data, capable of performing video tasks.
ii) LLaVA-NeXT (7B)~\cite{liu2024llavanext}, denoted as \capgenllavavideo, is a multi-modal large language model (MLLM) trained on instruction tuning data, capable of video tasks.
%
% iii) \ours{} using \texttt{LLaVA-Image-34B-Env}~\cite{liu2023llava}, an MLLM trained on larger instruction tuning data. 
iii) \ours{} using LLaVA (34B)~\cite{liu2023llava}, denoted as \capgenllavaimage, an MLLM trained on larger instruction tuning data. 
It is capable of image tasks only.
% \hg{The third and the fourth rows of \tabref{abl_arch} show the performance of each choice.}
%
Among these three caption generators,
% \texttt{LLaVA-Image-34B-Env}
\capgenllavaimage \ 
shows the best performance, despite being a zero-shot model and more sparsely applied.
Therefore, we employ
% \texttt{LLaVA-Image-34B-Env}
\capgenllavaimage \ 
as our default caption generator unless otherwise specified.

% \paragraph{Effect of infuser architecture.} 
% We compare various infuser architecture choices: addition (Add), cross-attention (CA), and concatenation (Concat.) (\ours{}). 
We also compare concatenation (Concat.), which we chose for our infuser architecture, with addition (Add) and cross-attention (CA). 
%
% \jw{Should say that details are in the appendix and put them in the appendix.}
Among these methods, concatenation shows the best performance.
For further details on the infuser architectures, refer to \appref{infuser_architecture}.
%
% \hg{We present the performance results in the last two rows of \tabref{abl_arch}.}
% Among these, concatenation shows the best performance.

\subsubsection{Quality of Environment Cues}
\begin{table}[t]
\centering
\resizebox{\columnwidth}{!}{
\begin{tabular}{l c cccc}
\toprule
& \textbf{R1@0.3} & \textbf{R5@0.3} & \textbf{R1@0.5} & \textbf{R5@0.5}
\\
\midrule
{\capgenllavaimage} (Ours)
% & \textbf{8.0} & \textbf{14.2} & \textbf{4.6} & 8.9 \\  % 버전: 102720
& \textbf{8.5} & \textbf{14.7} & 4.5 & \textbf{9.2} \\  % 버전: 102720
% EgoEnv~\cite{nagarajan2023egoenv}
EgoEnv
& 7.7 & 14.0 & \textbf{4.6} & 9.0 \\
% & 7.7 & 14.0 & \textbf{4.6} & \textbf{9.0} \\
\bottomrule
\end{tabular}
}
\caption{\tb{Ablation study}: quality of environment cues.}
\label{tab:abl_env_cues}
\end{table}

% \paragraph{Quality of environment cues.} 
% \hg{In \tabref{abl_env_cues}, }
\tabref{abl_env_cues} describes the ablation study of the quality of environment cues.
We compare using our environment cues with those from a prior work, EgoEnv~\cite{nagarajan2023egoenv}, which constructs a 3D environment through simulation to learn an \lfvlg{} model.
For a fair comparison, we use the same VLG model, VSLNet~\cite{zhang-etal-2020-span}, as used by EgoEnv.
Compared to EgoEnv, \ours{} demonstrates favorable performance, confirming the high quality of our environment cues even without 3D simulation.

\subsubsection{Environment Encoder}
\begin{table}[t]
\centering
\resizebox{\columnwidth}{!}{
\begin{tabular}{l c cccc}
\toprule
& \textbf{R1@0.3} & \textbf{R5@0.3} & \textbf{R1@0.5} & \textbf{R5@0.5}
\\
\midrule
{\ourtextenc~(Ours)} & 1.5 & \textbf{7.3} & 0.5 & 2.8 \\
% {\ours} & 1.8 & 7.6 & 0.8 & 3.0 \\  %        / all-MiniLM-L6-v2  / 성무진 교수님
% {\ourtextenc~(Ours)} & 1.5 & \textbf{7.7} & 0.4 & \textbf{2.8} \\  % 104454 / all-MiniLM-L6-v2  / 성무진 교수님 reproduce
\;\;SBERT $\rightarrow$ EgoVLP & \textbf{1.8} & \textbf{7.3} & \textbf{0.6} & \textbf{2.8} \\
% \;Loss: \\
\;\;MLL $\rightarrow \emptyset$ & 1.4 & 5.7 & 0.5 & 2.0 \\  % 생 all-mpnet-base-v2
% \;\;MLL $\rightarrow \emptyset$ & 1.0 & 4.7 & 0.4 & 1.5 \\  % 생 all-MiniLM-L6-v2
\;\;MLL $\rightarrow$ BCE & 1.5 & 6.4 & \textbf{0.6} & 2.5 \\  % 102720, all-mpnet-base-v2
% \;\;MLL $\rightarrow$ BCE & 1.4 & 7.1 & \textbf{0.6} & 2.7 \\  % 104404, all-MiniLM-L6-v2
% \;Text-model: \\
\bottomrule
\end{tabular}
}
\caption{\tb{Ablation study}: environment encoder.}
\label{tab:abl_text_encoder}
\end{table}

% \paragraph{Effect of environment encoder.} 
\tabref{abl_text_encoder} describes the ablation study of different training objectives and base text encoders.
We compare EgoVLP text model~\cite{linegocentric} and SentenceBERT~\cite{reimers-gurevych-2019-sentence} in a text only setting.
% \jw{Say that the details on the text only setting is described in the appendix.}
% \hg{
% For comprehensive details on the text only setting and the evaluation protocol, please check \appref{textonly}.
For details on the text only setting and the evaluation protocol, refer to \appref{textonly}.
% }
%
% \jw{Say that this is a text only setting.}
%
% Before edited by hyogun
% We choose SentenceBERT as our default text encoder as it shows favorable performance across all evaluation metrics.
% After edited by hyogun

% We choose SentenceBERT as our default text encoder because of its superior performance.
While ~\tabref{abl_text_encoder} shows the EgoVLP text model demonstrates favorable scores in a text-only setting, ~\tabref{abl_arch} shows it performs poorly when we integrate it into \ours{}.
Importantly, this discrepancy reveals the EgoVLP text model captures redundant information, which hinders the effective infusion of MLLM's experience into the VLG model.

% \paragraph{Effect of text encoder learning.} 
We study the effect of text encoder learning in a text only setting. 
Compared to using a frozen text encoder, fine-tuning the text encoder results in higher performance.
%
% \jw{Describe BCE loss briefly.}
% \jw{Put this in the appendix and say that the details on the BCE loss baseline is in the appendix.}
% \hg{
% We also train the model with BCE loss to compare the effectiveness of the MLL loss.
% BCE loss applies a sigmoid function $\sigma(x)=\frac{1}{1+\exp(-x)}$ instead of softmax to the dot products between the query and captions,
% obtaining $p_i = \sigma(\mathbf{z}_i^{\top} \mathbf{z}_q)$.
% If $y_i$ is $1$ if $i$ is within the GT interval $0$ otherwise, then BCE loss $L_\text{bce}$ is defined as follows:
% \[
% L_\text{bce}(\theta) = - \frac{1}{N} \sum_{i=1}^{N}
% y_i \log p_i + (1-y_i) \log (1 - p_i ) \ .
% \]
% }
%
% \sy{Please refer to the \secref{bce_loss}}
%
Compared to using BCE loss, using MLL loss in \eqnref{mll} show favorable performance.
Therefore, we employ MLL loss to fine-tune the text encoder by default.
Please refer to \appref{bce_loss} for the formal definition and details.

% % Before revision by Jinwoo
% % \subsection{Quantitative evaluation}
% % In this section, We compare the performance of our model with the current state-of-the-art model, on the EgoNLQ task. We show the results in  Table~\ref{tab:main_all}, {\ours} achieves the highest performance of an R5@0.3 score of 35.2\%, outperforming existing models GroundVQA~\cite{di2023groundvqa} that achieves R5@0.3 score of 34.3\%. This demonstrates the effectiveness of our model architecture and environment cues.
% % \hg{(why R5@0.3 is our primary metric) }

% \input{table/main_all}

% % After revision by Jinwoo
% \subsection{Comparison with State-of-the-Art}
% We compare the performance of the proposed method with the current state-of-the-art methods in \tabref{main_all}. 
% %
% {\ours} shows the best performance 3 out of 4 metrics and on par with state-of-the-art in terms of R1@0.3.
% %
% The results validate the effectiveness of using environmental cues provided by an MLLM as a proxy for human experiences.

% \subsection{Failure modes}  % 나중에

\

\section{Conclusion}
\label{sec:conclusions}

% Briefly summarize what we did in this work.

% Disclose all limitations

% Describe how potential future work can address these limitations and lead to interesting and ground breaking stuff.

% Before revision by Jinwoo
% In this work, we present a solution of long form video-language grounding by reducing search space. Our proposed method {\ours}, provide environmental cues through the Environment Encoder and Environment Infuser. The results show the effectiveness of {\ours}. We believe {\ours} will be advancements in long-term video understanding.

% After revision by Jinwoo
In this work, we present a solution to the long form video-language grounding problem.
Inspired by human capabilities of search space reduction, we propose {\ours}, infusing environmental captions extracted by a multi-modal large language model into a VLG model.
We validate the effectiveness of the proposed method on the challenging EgoNLQ dataset. 
The proposed method demonstrates favorable performance compared to existing methods, and we believe our contributions pave the way toward a new direction for addressing the long-form video understanding.

% Environment encoder, Environment Infuser를 통해 효과적으로 environment cues를 제공했다
% 여러 방법을 시도하여 더 발전된 environment cues를 생성하고자 한다.

% Before revision by Jinwoo
% \section*{Limitations}
% \ours{} offers academic advancements in solving VLG, but it has certain limitations.
% Firstly, the process of extracting captions using MLLM is GPU-intensive.
% Generating captions for 260 hours of videos requires 16 A5000 GPUs for 3.5 days (1.3k GPU hours).
% Consequently, we only apply \ours{} to a relatively small dataset,  EgoNLQ~\cite{grauman2022ego4d}.
% Future research can explore applying data-heavy methods such as NaQ~\cite{ramakrishnan2023naq}, which utilizes 3 million videos from Ego4D.
% Secondly, while MLLM operates faithfully on EgoNLQ, future work needs to verify its consistent performance across a wider variety of environments.

% After revision by Jinwoo
\section*{Limitations}
\label{sec:limitations}
Our work has some limitations.
The proposed method requires a caption generation process with an MLLM, which is computationally demanding.
For instance, generating captions for the EgoNLQ dataset, which is 260 hours long, requires approximately 1.3K GPU hours when using an NVIDIA RTX A5000 GPU.
In this work, the MLLM used performs reliably on the EgoNLQ dataset.
%
% However, we should verify its robust performance across a wider variety of datasets.
%
However, we should verify its robust performance across a wider variety of datasets, as it is possible that the MLLM used in this work may fail on certain datasets.
We plan to develop a solution for scenarios where the MLLM fails.

\section*{Ethical Considerations}

% \paragraph{Use of AI assistants.}
% We utilize ChatGPT-4o and GitHub Copilot to enhance the comprehensiveness and depth of our research.
% However, we ensure not to simply copy and paste, as this approach does not contribute to intellectual progress.

% \paragraph{Use of Ego4D dataset.}
The authors recognize that the Ego4D dataset~\cite{grauman2022ego4d} is proprietary, and intellectual property protected by copyright.
The authors who directly use Ego4D have agreed to the Ego4D license agreement, gaining access to the dataset.
This work only uses Ego4D for training and evaluation.
We will release the code and weights resulting from this research as open source.

The creators of the Ego4D dataset~\cite{grauman2022ego4d} have obtained informal agreements to distribute videos containing unblurred faces and have blurred all privacy-sensitive content.
Moreover, they note that the data collection protocol was reviewed and approved by University of Tokyo ethical review board~\cite{grauman2022ego4d}.
They have made extensive efforts to avoid ethical issues, though some may persist due to the large scale of the dataset.
In our use of the Ego4D dataset, we adhere to its ethical guidelines by:
\begin{itemize}
    \item Ensuring that all videos used for training and evaluation comply with the privacy standards set by the Ego4D creators.
    \item Not redistributing the dataset or embedding it visibly within our code.
\end{itemize}

\section*{Acknowledgment}

This work is supported by AI Center, CJ Corporation; by the Institute of Information \& Communications Technology Planning \& Evaluation (IITP) grant funded by the Korea Govenment (MSIT) (No. IITP-OTT: RS-2024-00353131); by the National Research Foundation of Korea (NRF) grant funded by the Korea government (MSIT) under (No. 2022R1F1A1070997); by the Artificial Intelligence Innovation Hub (No. RS-2021-II212068).

\bibliography{main}

% \newpage

\appendix

% \section{Example Appendix}
% \label{sec:appendix}

\section*{Appendix}

% \section{About EgoNLQ Dataset}\label{app:about_egonlq_dataset}
% EgoNLQ dataset consists of about 14k training samples and 4k validation samples, with an average video length of 8 minutes.

\section{Implementation Details}\label{app:implementational_detail}
In this section, we briefly provide our experimental setup and implementational details
\paragraph{Details of caption generator}
We use two versions of LLaVA, LLaVA-v1.6 (34B)~\cite{liu2023llava} for \capgenllavaimage \ and LLaVA-NeXT-Video-DPO (7B)~\cite{liu2024llavanext} for \capgenllavavideo \ as these provide environmental descriptions with only a single frame or an instant part of a video.

\paragraph{Data processing.}
For the input to LLaVA-v1.6 (34B)~\cite{liu2023llava}, we sample one frame every 10 seconds. For the input to LLaVA-NeXT-Video-DPO (7B)~\cite{liu2024llavanext}, we divide the entire video into 2-second clips and provided them as input. For the Video-Language Grounding Model and Environment Infuser, we used concatenated features from EgoVLP~\cite{linegocentric} and InternVideo~\cite{chen2022ego4d}.

\paragraph{Model training.}\label{sec:model_training}
For training the Video-Language Grounding Model and Environment Infuser, we use 8 A5000 GPUs and the AdamW optimizer with a learning rate of 1e-5.
We trained the model for 20 epochs.
The overall architecture of \ours{} has 231M learnable parameters in total.  % 108M(SBERT) + 123M(EI + VLG model)

\paragraph{Pre-trained weights.}
We use SentenceBERT with \texttt{all-mpnet-base-v2} for the environment encoder. We also use the GroundVQA~\cite{di2023groundvqa} weights pre-trained on the EgoNLQ dataset for training the Video-Language Grounding Model and Environment Infuser.

\section{Infuser Architecture Choices}\label{app:infuser_architecture}
We conduct with various structures to configure the Environment Infuser: i) The Add architecture directly adds environment cues to the video features. ii) The Cross-attention architecture passes through a cross-attention layer between the environment cues and video features. iii) Concatenation architecture concatenates the environment cues and video features, followed by passing through a linear projection
% \sy{do we need figures to explain architecture?}

\begin{figure}[t]
\centering
    \includegraphics[width=\linewidth]{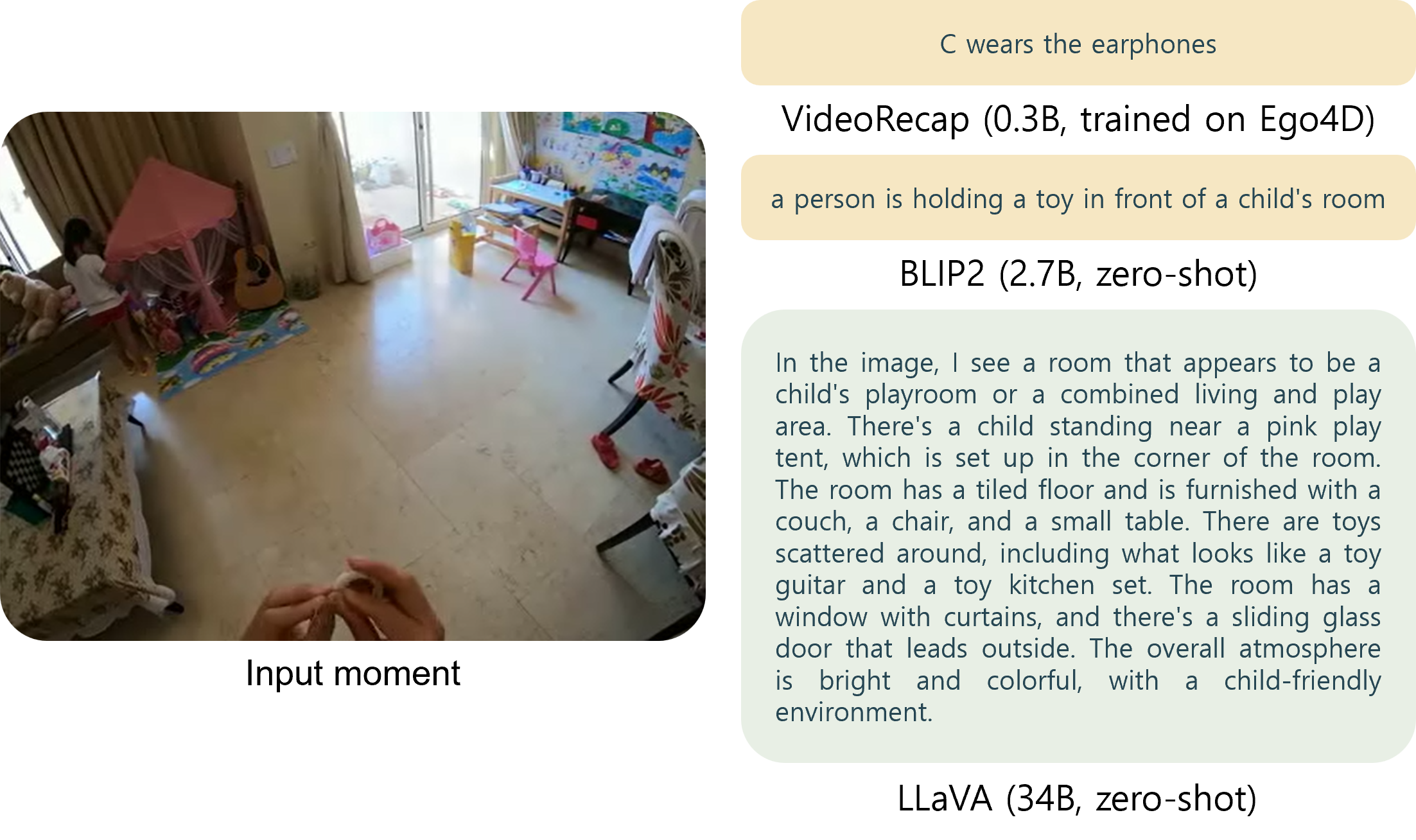}
    \figcaption{We should use a large caption generator}{
    We need fine-grained descriptions to reduce the search space within a long sequence of indistinguishable in-context views.
    Not only where I am, but also the direction I see, objects, and their relative locations.
    }
    \label{fig:llava}
\end{figure}

\section{Text-only VLG Protocol}\label{app:textonly}
We demonstrate a protocol to obtain an interval prediction from query-caption similarity without video to roughly measure VLG performance.
This metric helps gauge how well the text encoder extracts fine-grained information beneficial to VLG from MLLM's rich environmental descriptions. 

To measure the performance, we need to obtain the start and the end indices of the moment prediction from query-caption similarities.
Given the text encoder's encoded feature $\mathbf{Z}_\text{e}$ with the $i$-th vector $\mathbf{z}_i \in \mathbb{R}^{D_t}$ and a query feature $\mathbf{z}_q \in \mathbb{R}^{D_t}$,
for a caption index set $I = \{ 1, 2, \dots, N \}$,
we obtain the caption index $i^*$ with the highest dot product value between caption and query:
\[ i^* = \argmax_{i \in I} \mathbf{z}_i^{\top} \mathbf{z}_q \ .\]

Among $M$ frames, we find the index of the closest one:
\[ j^* = \left\lfloor \frac{i^*}{N} M + \frac{1}{2} \right\rfloor \in \{ 1, 2, \dots, M \} \ . \]
For a predefined span width $M_\text{span}$, the start and end indices of the text-only predicted interval $\hat{s}, \hat{e}$ are:
\[
\hat{s} = \left\lfloor j^* - \frac{M_\text{span}}{2} \right\rfloor, \quad \hat{e} = \left\lceil j^* + \frac{M_\text{span}}{2} \right\rceil \ .
\]

We set $M_\text{span}$ to correspond to 30 seconds.

\section{BCE Loss for Text Encoder Learning}\label{app:bce_loss}
We also train the model with BCE loss to compare the effectiveness of the MLL loss.
BCE loss applies a sigmoid function $\sigma(x)=\frac{1}{1+\exp(-x)}$ instead of softmax to the dot products between the query and captions,
obtaining $p_i = \sigma(\mathbf{z}_i^{\top} \mathbf{z}_q)$.
If $y_i$ is $1$ if $i$ is within the GT interval $0$ otherwise, then BCE loss $L_\text{bce}$ is defined as follows:
\[
L_\text{bce}(\theta) = - \frac{1}{N} \sum_{i=1}^{N}
y_i \log p_i + (1-y_i) \log (1 - p_i ) \ .
\]

\section{Loss for Video-Language Grounding}\label{app:vlg_loss}
We employ two different forms of VLG loss ($L_\text{vlg}$) depending on the choice of the temporal localization head.

For VSLNet~\cite{zhang-etal-2020-span} head, we utilize the query-guided highlighting (QGH) loss $L_\text{QGH}$ and the span loss $L_\text{span}$ , following the original work.
We define the VLG loss as the sum of these two losses, resulting in $L_\text{vlg} = L_\text{QGH} + L_\text{span}$.
We use this form for the models in \tabref{abl_env_cues}.

For ActionFormer~\cite{zhang2022actionformer} head, we adopt the binary focal loss $L_\text{focal}$~\cite{Lin_2017_ICCV} and the DIoU loss $L_\text{DIoU}$~\cite{zheng2020diou}.
Again, we simply define the VLG loss as the sum of these two losses: $L_\text{vlg} = L_\text{focal} + L_\text{DIoU}$.
We train the models in \tabref{main_all}, \tabref{abl_arch} with this form.

% \section{Faithfulness and Completeness of LLaVA Captions}

% This is an appendix.

% \newpage
% \input{999_rebuttal}

\end{document}